\documentclass[acmtog]{acmart}
\acmSubmissionID{335}
\usepackage{amsmath}
\usepackage{booktabs} 

\usepackage{xcolor}
\usepackage{graphicx}
\usepackage{graphbox}
\usepackage{array}
\usepackage{enumitem}
\usepackage{xspace}

\definecolor{rowblue}{RGB}{220,230,240}
\definecolor{myorchid}{RGB}{150,10,30}
\definecolor{myblue}{RGB}{10,30,250}
\definecolor{mygreen}{RGB}{10,190,10}
\definecolor{myred}{RGB}{190,20,20}

\newcommand{\linkk}[1]{\emph{\textcolor{magenta}{#1}}}

\newcommand{\mytilde}{\raise.17ex\hbox{$\scriptstyle\mathtt{\sim}$}}

\newcommand{\update}[1]{{#1}}

\newcommand{\supp}[1]{#1}

\newcommand{\vvv}[1]{\mathbf{#1}}
\newcommand{\norm}[1]{\left\lVert#1\right\rVert}

\usepackage[ruled]{algorithm2e} 

\SetAlFnt{\small}
\SetAlCapFnt{\small}
\SetAlCapNameFnt{\small}
\SetAlCapHSkip{0pt}

\AtBeginDocument{%
 \abovedisplayskip=6pt plus 3pt minus 5pt
 \abovedisplayshortskip=0pt plus 3pt
 \belowdisplayskip=6pt plus 3pt minus 5pt
 \belowdisplayshortskip=5pt plus 3pt minus 4pt
}

\makeatletter
\renewcommand{\paragraph}{%
  \@startsection{paragraph}{4}%
  {\z@}{0.7ex \@plus 1ex \@minus .1ex}{-1em}%
  {\normalfont\normalsize\bfseries}%
}
\makeatother

\makeatletter
\@namedef{ver@everyshi.sty}{} 
\makeatother

\acmJournal{TOG}

\setcopyright{acmcopyright}\acmJournal{TOG}
\acmYear{2021}\acmVolume{40}\acmNumber{6}\acmArticle{1}\acmMonth{12} \acmDOI{10.1145/3478513.3480546}
\begin{document}
\title{Layered Neural Atlases for Consistent Video Editing }

\author{Yoni Kasten}
\affiliation{
  \institution{Weizmann Institute of Science}
}
\email{yonikasten@gmail.com}\author{Dolev Ofri}
\affiliation{
  \institution{Weizmann Institute of Science}
 }
\email{dolev.ofri@weizmann.ac.il}
\author{Oliver Wang}
\affiliation{
  \institution{Adobe Research}
}
\email{owang@adobe.com}
\author{Tali Dekel}
\affiliation{
  \institution{Weizmann Institute of Science}
  }
\email{tali.dekel@weizmann.ac.il}

\begin{teaserfigure}
\centering
\includegraphics[width=.95\textwidth]{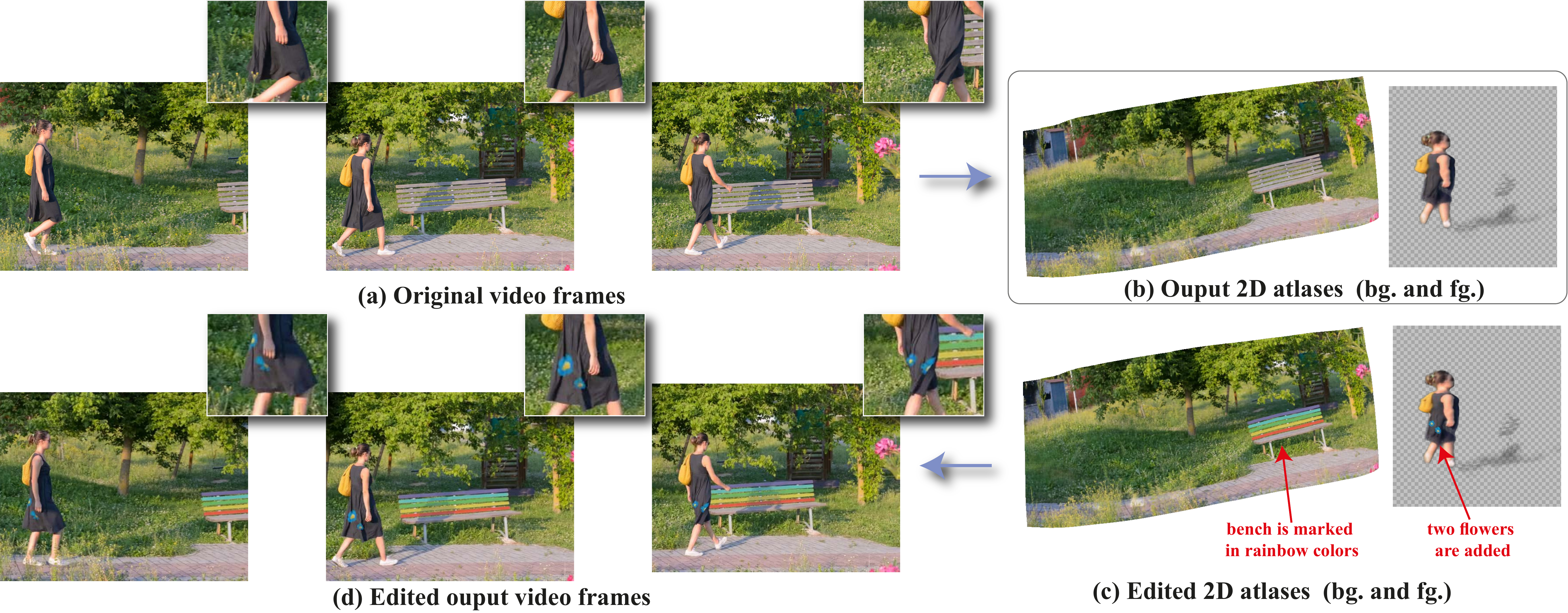}
\caption{\label{fig:teaser} Our method enables \textit{temporally consistent} and easy video editing via manipulation in the 2D atlas domain. Given a natural video as input (a), our method estimates a set of \emph{layered 2D atlases} (b), each providing a unified and interpretable parameterization of an object's, or background's appearance throughout the video. Our method further estimates a mapping from each pixel in the video to a 2D point in each atlas, and its opacity. This enables any edits made to the atlases (c) to be consistently and automatically propagated to all frames of the video (d). In this example, edits have been applied to the two atlases: flowers are added to the dress, and the bench is painted with rainbow colors; these edits are faithfully projected to the original frames, while preserving the deformation of the dress and complex scene effects as shadows. \supp{Please see the supplementary material (SM) for video results.}} 
\end{teaserfigure}

\begin{abstract}

We present a method that decomposes, and ``unwraps'', an input video into a set of \emph{layered 2D atlases}, each providing a \emph{unified representation} of the appearance of an object (or background) over the video.
For each pixel in the video, our method estimates its corresponding 2D coordinate in each of the atlases, giving us a consistent parameterization of the video, along with an associated alpha (opacity) value.
Importantly, we design our atlases to be interpretable and semantic, which facilitates easy and intuitive editing in the atlas domain, with minimal manual work required.
Edits applied to \emph{a single 2D atlas} (or input video frame) are automatically and consistently mapped back to the original video frames, while preserving occlusions, deformation, and other complex scene effects such as shadows and reflections.
Our method employs a coordinate-based \update{Multilayer Perceptron (MLP)} representation for mappings, atlases, and alphas, which are jointly optimized on a per-video basis, using a combination of video reconstruction and regularization losses.
By operating purely in 2D, our method does not require any prior 3D knowledge about scene geometry or camera poses, and can handle complex dynamic real world videos.
We demonstrate various video editing applications, including texture mapping, video style transfer, image-to-video texture transfer, and segmentation/labeling propagation, all automatically produced by editing a single 2D atlas image. Project page: \href{https://layered-neural-atlases.github.io/}{ \linkk{https://layered-neural-atlases.github.io/}}
\end{abstract}

%
%


%
%


\maketitle

\section{Introduction}

While \emph{image} editing and manipulation tools have seen steady progress, allowing complex editing effects to be achieved by novice users, \emph{video} editing remains a difficult task that is largely restricted to the domain of professionals or semi-professionals. 
This is because editing video content poses two key additional challenges.
First, edits need to be applied in a temporally consistent manner to all frames.
For example, a texture mapped onto a moving object must move consistently with the object, and occlude or be occluded by the scene correctly.
Second, editing interfaces need to be able to represent temporal content in an intuitive manner. 
While we can easily interact with a single frame, managing a spatio-temporal volume of pixels is significantly more challenging. 
This raises the question, \emph{how can we represent video content in order to facilitate intuitive and temporally consistent editing?}

The above challenges have been tackled by both 2D and 3D approaches.
In 2D approaches, users edit keyframes from the video and then the edit is propagated to the rest of the frames using frame-to-frame tracking~\cite{ae} or feature similarity~\cite{jabri2020space}.
These approaches tend to suffer from drift and fail at occlusions, and the edit experience is highly dependent on the selection of keyframes. 
In 3D approaches, a geometric model of the scene is estimated, edits are then applied in 3D, and mapped back to the video frames.
This approach is most commonly used in a domain where strong 3D priors exist (for example, face filters in Snapchat), or on static scenes (e.g., \cite{luo2020consistent, xiang2021neutex}).

We take a different route and propose a novel 2D, layer-based method that enables \emph{consistent} editing of natural videos, containing arbitrary types of moving objects or background.
We draw inspiration from pre-deep learning video mosaic approaches, as well as recent deep learning methods for neural video layering, and propose to represent the video via a set of \emph{layered neural 2D atlases}, one for each object to be edited, and one for the rest of the content (background). 
Each of our atlases serves as a \emph{unified representation} of an object's or background's appearance over the video (see Fig.~\ref{fig:teaser}(b)).
Specifically, for each pixel in the video, our method computes its corresponding 2D point in each of the atlases, and determines the visibility of each matched atlas point via an associated alpha (opacity) value.
With this mapping in hand, we can parameterize the original video through the set of 2D atlases, thereby allowing us to train the model using the input video as self-supervision. 
As our method operates entirely in 2D, it  can handle diverse real-world videos with complex dynamic scenes, where estimating 3D geometry would be an extremely challenging task.

Our method employs an MLP-based, continuous representation for all our components: the mappings, atlases, and alphas.
All networks are jointly optimized on a per-video basis, using a combination of video reconstruction and regularization losses.
Our goal is to reconstruct atlases that serve as a \emph{visual proxy} of the video for intuitive editing.
To this end, we design our atlases to be semantic and interpretable by encouraging the mapping from video to atlas to be locally rigid and consistent, i.e., corresponding 2D pixels in consecutive frames should be mapped to the same atlas point.
Consequently, editing can be applied directly in the atlas domain by simply editing  \emph{a single 2D atlas} (or input video frame).
The edits are then automatically and consistently mapped back to the original video frames, while preserving occlusions, deformation, and other complex scene effects such as shadows and reflections (see Fig.~\ref{fig:teaser}). 
We apply this framework for various video editing applications, including texture mapping, video style transfer, image-to-video texture transfer, and segmentation/labeling propagation.

Our design choices have several advantages: (i) our atlas space is \emph{continuous}, fully differentiable, and flexible.
Specifically, we avoid the need to discretize the atlas to a fixed resolution during training, allowing the network to distribute the content in 2D space as it likes in order to minimize the objective.
Consequently, our MLP-based atlases lead to better visual quality and better decomposition than a standard fixed pixel grid atlas, as demonstrated by our experiments. 
(ii) Unlike prior work that makes the assumption of a static camera or background, we allow the mapping network to deform space, which leads to atlases that can model complex geometries, as well as temporal effects like parallax and pose changes, while remaining interpretable to users. 
(iii) Finally, our layered decomposition and mapping allows us to composite the user edits directly with the original video frames, which makes our method applicable even in cases where the accurate reconstruction of \emph{all} video elements in a scene is challenging. 

In summary, we present the following contributions: 1) We introduce an end-to-end coordinate-MLP based framework that decomposes and maps a video into a set of layered 2D atlases. 2) We carefully design several regularization terms that encourage atlas interpretability and facilitates easy and consistent video editing via 2D atlas manipulation, and 3) we demonstrate a variety of video editing applications that this decomposition enables on complex real-world videos with deformation, reflections, and occlusion.

\section{Prior Work}

\paragraph{Video mosaics.} 
Stitching panoramas from multiple images is a fundamental and long standing task in computer vision~\cite{brown2007automatic}.
Here, we focus only on methods related to creating mosaics from dynamic video. 
One application of video mosaics is as a convenient way to summarize the content of the video. 
Work on video indexing builds mosaics using homography warping for compression~\cite{irani1995mosaic}, or video summarization~\cite{irani1998video}.
Follow up work composites multiple instances of foreground objects onto a background mosaic~\cite{correa2010dynamic}, or uses patch-based copying to smoothly blend dynamic content from a video into a tapestry~\cite{barnes2010video} in order to visualize dynamic content in a single image.
In panoramic video textures~\cite{agarwala2005panoramic} the goal is to build an animated panorama for stochastic motion. 
Homography based panorama construction has also been used as a way to convert 2D to 3D-stereo videos, where the background fills in holes that appear by popping out foreground edits~\cite{schnyder20112d,ribera2012video}.
Our motivation is different -- our main purpose is to reconstruct atlases for enabling temporally consistent and easy editing of natural videos.
Furthermore, most of these methods rely on a parametric alignment model for the background (e.g, via homographies), and are thus restricted to rotation-only camera motion, or a planar scene.
In contrast, our method allows for the mapping from video to atlas to be nonrigid, and hence can handle camera motion with parallax.

\begin{figure*}[t]
    \centering
    \includegraphics[width=.95\linewidth]{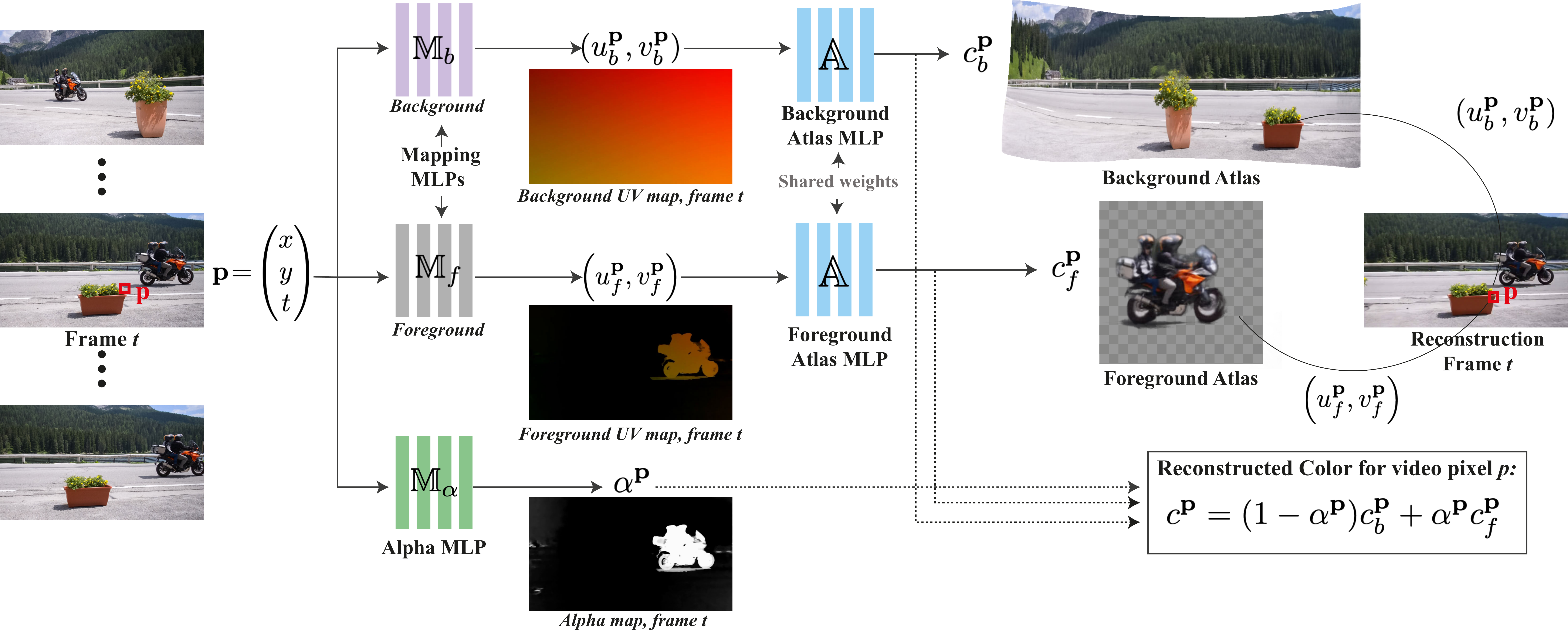}
    \caption{\textbf{Illustration of our pipeline for two atlases:} Each pixel location in the video $p$ is fed into two mapping MLPs, $\mathbb{M}_b$, $\mathbb{M}_f$, which predict the corresponding 2D $\update{(}u,v\update{)}$ coordinate of $p$ in each atlas. These coordinates are then fed into an atlas MLP $\mathbb{A}$, which outputs the RGB color at the that location (the foreground and background atlases are mapped into two different regions in the 2D atlas space). The visibility of $p$ in each atlas is determined by an alpha MLP, $\mathbb{M}_a$, which takes $p$ as input and predicts an opacity value. The RGB color at $p$ can then be reconstructed by alpha-blending the predicted atlas points. All networks are trained end-to-end where the main loss is a self-supervised reconstruction loss of the original input video. For visualization proposes, we show the predicted mappings and alpha map for a given frame in the video, and render the atlases as an RGB image.  
    \label{fig:pipeline}}
\end{figure*}

Our work is inspired by Unwrap Mosaics~\cite{rav2008unwrap}, which decomposes a video into a set of 2D mosaics, and computes a mapping from each mosaic to each video frame.
Editing is then applied directly on the mosaics, and mapped back to the video. 
We share the same editing goal, but our method differs in a few key ways: (i) \emph{alpha-layering:} they break the video using a set of pre-computed, binary segmentation masks, hence cannot represent any semi-transparent effects (e.g., reflections or shadows), and cannot fix any errors in the input masks during optimization. Our method starts with rough binary masks but then refines them with the final reconstruction objective generating time-varying alpha mattes per atlas.  (ii) \emph{tracking:} they rely on feature tracking to initialize their mosaic by embedding the tracks into one global 2D coordinate system. Establishing trajectories in natural, complex videos is prone to errors due to lighting changes and non rigid object motion. In contrast, our method uses only \emph{local} motion estimates (optical flow) as a \emph{soft constraint}. 
As such, we show that we can handle complex dynamic scenes for which accurate feature tracking cannot be easily estimated (e.g. water). 
(iii) \emph{discretization and optimization:}
in their method, the mosaic is represented as a discrete image whose resolution has to be fixed.
Our neural representation for the atlases and mappings functions allows us to work in a \emph{continuous} atlas domain, hence avoid discretization during training (see Sec.~\ref{sec:gridatlas}). 
Finally, unlike \cite{rav2008unwrap} which uses a complex, alternating  optimization framework, our networks are jointly optimized via simple, standard gradient decent optimization.

\paragraph{Video layer decomposition.}
Our method represents a video with a set of \emph{layered} 2D atlases. 
Traditional workflows for selective editing have long used various forms of layer decomposition~\cite{wang1994representing,shade1998layered}.
Most common, layers (masks) are computed by extensive manual work to rotoscope and select regions, or are based on color similarity~\cite{lin2017layerbuilder}.
Recently, Lu et al.~\shortcite{lu2020layered, lu2021omnimatte} propose using neural networks to decompose a video into a set of color and opacity layers.
Their framework takes a video with rough object segmentation masks, and trains a CNN to output RGBA (color image + alpha matte) layers such that when composited, yields the original video content.
A key difference between this line of work and ours is that they provide a \emph{per-frame} representation, i.e., a set of RGBA layers for each frame of the video, and hence video editing is restricted to directly manipulating frame layers (e.g., removing/retiming a layer throughout the video).
This approach cannot support consistent propagation of edits \emph{across} time, which is our primary goal.
In addition, prior work~\cite{lu2020layered, lu2021omnimatte} rely on the assumption that the background of each frame can be aligned using homographies, while our method allows for flexible mappings into atlases, enabling our approach to work on cases with moving cameras or background.

\paragraph{Video propagation approaches.}
A natural alternative approach to consistent video editing is to propagate edits from one frame to the next.
Video Propagation Networks~\cite{jampani2017video} splat edits into a bilateral space, and refine them to subsequent frames. 
Alternate approaches have used phase-based edits~\cite{meyer2016phase}, or optical flow~\cite{texler2020interactive} to propagate key-frame based edits in video.
These approaches work well when correspondences can be established, but have challenges at occlusions and motion discontinuities.
Recently, deep learning based solutions have been proposed to help overcome shape change and problems with occlusions, specifically for tasks like mask and label propagation~\cite{oh2018fast}.
Several methods have proposed to use self supervised learning to learn features that can then be used for video label propagation, formulated as a cross-time attention operation~\cite{wang2019learning,jabri2020space}.
Unlike these methods, we explicitly reconstruct an RGB atlas, which represents content over \emph{multiple frames}, allowing for intuitive editing of content beyond single keyframes.
In addition, we compare our result to this category and show that our method generates a smoother parameterization over time, which is more suitable for the task of video editing.

\paragraph{Continuous image/video representations.}
Our method follows a trend in recent work to use coordinate-based MLPs to represent image and video content. 
Such methods have shown success in being able to represent 3D geometry~\cite{Park_2019_CVPR,groueix2018papier,mescheder2019occupancy,mildenhall2020nerf}, and dynamic video scenes~\cite{li2020neural}.
Follow up works such as  \update{\cite{tancik2020fourfeat} and} SIRENs~\cite{sitzmann2020implicit} showed that such MLPs can be used to overfit image and video content with high accuracy using \update{respectively fourier features maps or} sinusoidal activation layers.
In a similar vein, Chen et al.~\shortcite{chen2020learning} propose to use MLPs and a feature grid for image super resolution.
In our work, we are motivated by these advances, and use similar coordinate-based MLPs for the task of atlas decomposition and video editing.

\section{Estimating Neural Atlases from Video}

The input to our method is a natural video of a dynamic scene, and a coarse mask identifying the object(s) of interest.
Our goal is to estimate: (1) a set of 2D atlases, one for the background and one per dynamic object of interest, (2) a mapping from each video pixel to a 2D coordinate in each atlas, and (3) opacity values at each pixel w.r.t. each atlas.

We employ coordinate-based MLPs to represent each component, which enables us to work in a \emph{continuous} space and avoid discretization of both the atlases and mappings during training.
For editing purposes we render (discretize) the atlas into a fixed image grid; the user edits the rendered atlas image, and the edits are then automatically and consistently mapped back to the video frames using the mapping networks. 

For the majority of our experiments, we focus on the common case of a single dominate moving object, hence two atlas layers, one for the background and one of the moving object.
However, our method can be applied on scenes with multiple foreground objects and atlases (see example in Sec.~\ref{sec:multiple_object}).
Our two layer framework is illustrated in Fig.~\ref{fig:pipeline}, and consists of four MLP networks (\supp{Please see Appendix~\ref{app:arch} for the exact architecture used}).
First, the mapping networks ($\mathbb{M}_b, \mathbb{M}_f$) take a pixel location as input and output its corresponding 2D point in each atlas.
The predicted 2D coordinates are then fed to the atlas network ($\mathbb{A}$) that outputs the atlas' RGB color at that location.
Each pixel location is also fed into the Alpha MLP ($\mathbb{M}_\alpha$) which outputs the opacity of each atlas at that location.
The initial decomposition of the layers is achieved by bootstrapping the Alpha network using rough object masks that can be automatically computed (e.g., using MaskRCNN) or provided by the user.
With these networks in hand, the reconstructed RGB color at each video pixel is estimated by alpha-blending the corresponding atlas colors. 

\update{We assume that object's motion is consistent over time and that there are no significant self-occlusions, in which case the object can be reliably represented via a single atlas layer. However, if the object undergoes complex non-rigid motion and self-occlusions, we opt to decompose the object into multiple atlas layers. To this end, we provide a simple interface for the user to annotate rough scribbles that are used to to guide the object's decomposition (Sec.~\ref{sec:user_input}). 

}

\subsection{Training}
\label{section:training}

Our entire framework is trained end-to-end, in a self-supervised manner, where the main loss is a reconstruction loss to the original video.
However, a reconstruction loss alone is insufficient for producing meaningful and semantic atlases that can be used for editing \update{(Sec.~\ref{sec:ablation_res})}. 
To this end, we design several regularization losses on the mapping and decomposition:
\begin{enumerate}
    \item \emph{Rigidity loss:} we preserve the local structure of objects as they appear in the input video by encouraging the mapping from video pixels to atlas to be locally rigid.
    \item \emph{Consistency loss:} we encourage corresponding pixels in consecutive frames of the video to be mapped to the same atlas point; pixel correspondence is computed using an off-the-shelf optical flow method~\cite{teed2020raft}.
    \item \emph{Sparsity loss:} we encourage the networks to recover the minimal content needed to recover the video in atlases via a sparsity loss. 
    
\end{enumerate}
Therefore, our total loss is given by:
\begin{equation}
\mathcal{L}=\mathcal{L}_{\textit{color}}+\mathcal{L}_{\textit{rigid}}+\mathcal{L}_{\textit{flow}}+\mathcal{L}_{\textit{sparsity}}.
\end{equation}

\begin{figure}[t]
    \centering
    \includegraphics[width=.4\textwidth]{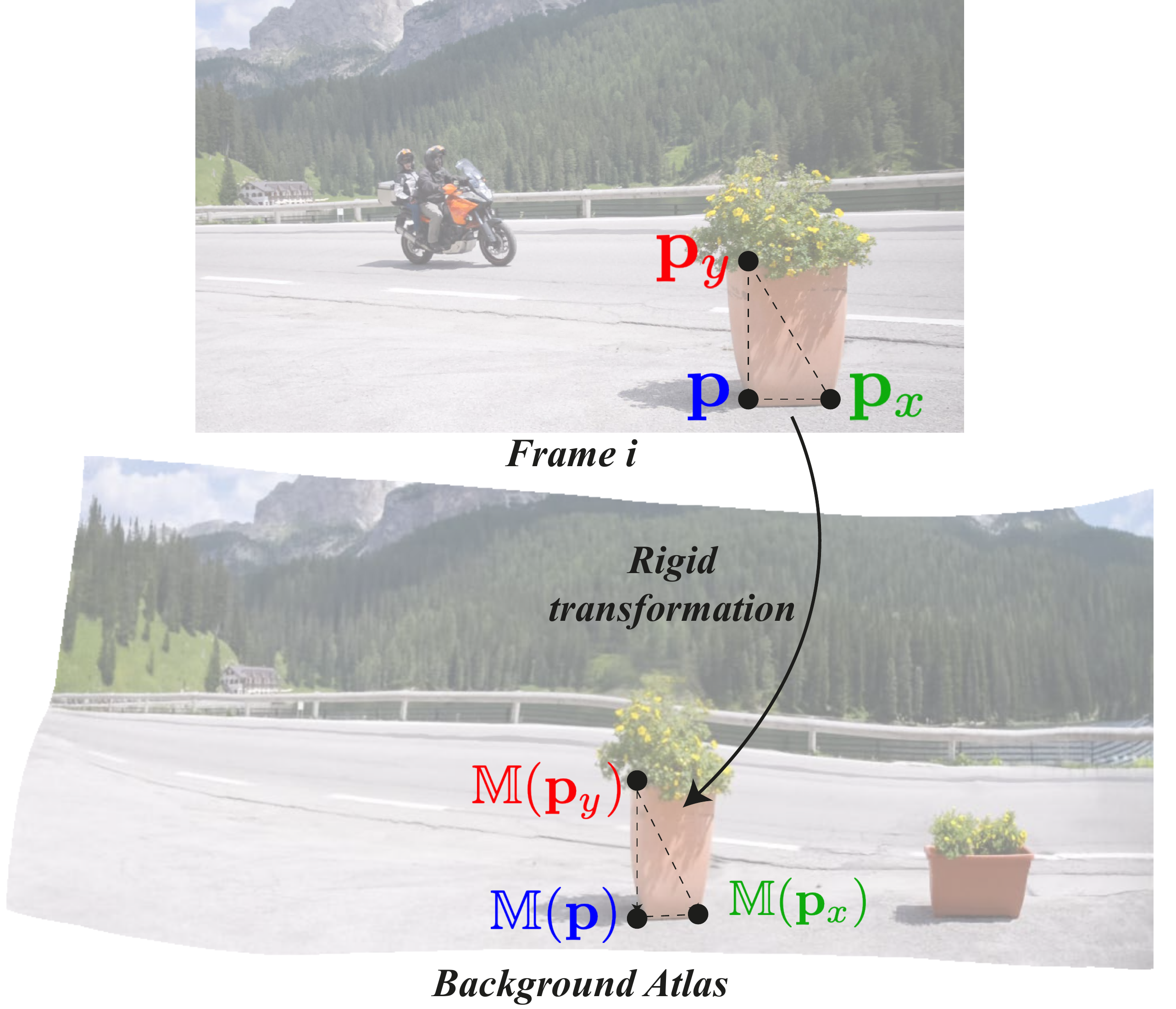}
    \caption{{\bf Rigidity loss illustration:} Our rigidity loss is defined over a tuple of 2D points: a pixel $p$, and horizontal and vertical offsets pixels ($p_x$ and $p_y$). We encourage a rigid transformation between the resulting triangle in the video frame (top), and the mapped triangle in the atlas (bottom). See Sec.~\ref{sec:losses} for formal definition. \vspace{-2mm}}
    \label{fig:losses}
\end{figure}

\subsection{Model Overview} \label{section:model_overview}
Each pixel location $\vvv{p}=(x,y,t)^T\in \mathbb{R}^3 $, is fed into two mapping networks: \begin{equation*}
    \mathbb{M}_f :\mathbb{R}^3\rightarrow \mathbb{R}^2,~~\mathbb{M}_b: \mathbb{R}^3\rightarrow \mathbb{R}^2
\end{equation*} where $\mathbb{M}_f$, $\mathbb{M}_b$ map $\vvv{p}$ to a 2D point in the foreground and background atlas regions respectively:
\begin{equation}
    \begin{array}{c}
    (u_b^{\vvv{p}}, v_b^{\vvv{p}}) = \mathbb{M}_b(\vvv{p}) \quad \text{and} \quad (u_f^{\vvv{p}}, v_f^{\vvv{p}}) = \mathbb{M}_f(\vvv{p})
    \end{array}
\end{equation}
We assume that the mapping to each atlas is smooth, which we implicitly impose by \emph{not} applying positional encoding~\cite{mildenhall2020nerf} to $\vvv{p}$ in $\mathbb{M}_f,\mathbb{M}_b$.

While one could use separate networks $\mathbb{A}_f,\mathbb{A}_b$ to represent foreground and background atlases, we found it sufficient to use a single atlas $\mathbb{A}$, and restrict mapping networks $\mathbb{M}_f$ and $\mathbb{M}_b$ to point into separate pre-defined quadrants in continuous [-1,1] space.
The atlas $\mathbb{A}$ then predicts the corresponding $(R,G,B)$ color for any given $\update{(}u,v\update{)}$ coordinate.
As we desire our atlas to represent high frequency appearance information such as edges and texture, we first pass the 2D atlas coordinates through positional encoding~\cite{mildenhall2020nerf}, denoted by $\phi(\cdot)$. The predicted colors are then provided by: 
\begin{equation}
    \begin{array}{c}
\vvv{c}_b^{\vvv{p}} = \mathbb{A}\left(\phi(u_b^{\vvv{p}}), \phi(v_b^{\vvv{p}})\right), \\ \\
\vvv{c}_f^{\vvv{p}} = \mathbb{A}\left(\phi(u_f^{\vvv{p}}), \phi(v_f^{\vvv{p}})\right)
    \end{array}
    \label{eq:colors}
\end{equation}
where $\mathbb{A} :\mathbb{R}^{4N}\rightarrow [0,1]^3$ is the Atlas MLP network, and $N$ is the number of frequencies used in the positional encoding. 
Each positionally encoded pixel location is also fed into $\mathbb{M}_\alpha :\mathbb{R}^{6N}\rightarrow [0,1]$ that predicts $\alpha^{\vvv{p}} = \mathbb{M}_\alpha (\phi(\vvv{p}))$, the opacity of the foreground atlas layer in pixel location $\vvv{p}$. 
The RGB color at $\vvv{p}$ can be then reconstructed by alpha-blending the corresponding atlas points from each layer:
\begin{equation}\label{eq::pixel_reconstruction}
    \vvv{c}^{\vvv{p}} = (1 - \alpha^{\vvv{p}}) \vvv{c}^{\vvv{p}}_b +  \alpha^{\vvv{p}} \vvv{c}^{\vvv{p}}_f
\end{equation}

\subsection{Loss terms}
\label{section:losses}
\label{sec:losses}
\paragraph{Reconstruction loss.}
 Our reconstruction loss is the main loss in our objective. Formally, it is given by:
\begin{equation}
    \mathcal{L}_{\textit{color}}= \beta_r E_{\textit{RGB}}+\beta_g E_{\textit{Grad}}
\end{equation}
The first term is the squared distance between the ground truth color, $\bar{c}^\vvv{p}$,  and  the inferred color at $c^\vvv{p}$, according to Eq.~\ref{eq::pixel_reconstruction}:
\begin{equation}
    E_{\textit{RGB}}=\norm{\vvv{\bar{c}}^\vvv{p}-\vvv{c}^\vvv{p}}_2^2,
\end{equation}
and the second term is a loss on the image gradients:
\begin{equation}
    E_{\textit{Grad}}=\norm{\vvv{\bar{\nabla}_x}-\vvv{{\nabla}_x}}_2^2+\norm{\vvv{\bar{\nabla}_y}-\vvv{{\nabla}_y}}_2^2
\end{equation}
where $\{\bar{\nabla}_x, \bar{\nabla}_y\}$ are the ground truth spatial derivatives at $\vvv{p}$, and $\{\nabla_x, \nabla_y\}$ are the computed spatial derivatives using our prediction, and $\beta$s are weights.

\paragraph{Rigidity loss}
Minimizing the reconstruction loss alone can result in a solution where the atlas can not be easily edited.
One extreme example of this would be that the atlas could be simply a color palette and the mapping a lookup function into it. 
To avoid this, we propose to keep the atlas as undistorted as possible (with respect to the original video frames), while still allowing \emph{some} distortion to handle object non-planarity and other effects. 
To this end, we encourage the mapping from pixel location in the video to the 2D atlas to be locally rigid (see Fig.~\ref{fig:losses}). Formally, we define our local rigidity loss through the Jacobian matrix of the mapping $\mathbb{M}$ at $\vvv{p}$, which  is defined by:   
\begin{equation}
    J_\mathbb{M}=\begin{bmatrix} \mathbb{M}(\vvv{p}_x)-\mathbb{M}(\vvv{p})  & \mathbb{M}(\vvv{p}_y)-\mathbb{M}(\vvv{p}) \end{bmatrix}\in \mathbb{R}^{2\times 2}
\end{equation}
We approximate the Jacobian computation using $\vvv{p}_x=(x+\Delta,y,t) ,\vvv{p}_y=(x,y+\Delta,t)$ where $\Delta$ corresponds to an offset in pixels (here $\Delta=1$).
Let $\sigma_1,\sigma_2$ be the two singular values of $J$. For a rigid mapping, it holds that $\sigma_1=\sigma_2=1$, which is what we want to encourage. We do so by using  a variant of symmetric Dirichlet term~\cite{rabinovich2017scalable} and define the rigidity term for a given mapping $\mathbb{M}$,  at pixel $\vvv{p}$ by:
\begin{equation}
    E_{\textit{Rigid}}(\mathbb{M})=\norm{J^{\small T}_{\mathbb{M}}J_{\mathbb{M}}}_F+\norm{\left(J_{\mathbb{M}}^TJ_{\mathbb{M}}\right)^{-1}}_F
\end{equation}

We apply the above term to both foreground and background mapping, resulting in: 
\begin{equation}
    \mathcal{L}_{\textit{rigid}}= \beta_r(E_{\textit{Rigid}}( \mathbb{M}_b)+ E_{\textit{Rigid}}( \mathbb{M}_f))
\end{equation}

\paragraph{Optical flow loss.} 
Ideally, we want each point in the scene to be mapped consistently to the same atlas point.
While the network can satisfy this to some degree, we found that adding an explicit regularization improved the results.
Specifically, we encourage a consistent mapping by defining the following loss:
\begin{equation}
    E_{\textit{Flow}}=\alpha^\vvv{p} \norm{ \mathbb{M}_f(\vvv{p})-\mathbb{M}_f(\vvv{p'})} +  \left(1-\alpha^\vvv{p}\right) \norm{ \mathbb{M}_b(\vvv{p})-\mathbb{M}_b(\vvv{p'})}
\end{equation}
where $\vvv{p'}$ is the corresponding point of $\vvv{p}$ at frame $t\pm1$ as determined by \update{pre-computed} optical flow.
Similarly, we expect corresponding pixels to have the same alpha value:
\begin{equation}
     E_{\textit{Flow}-\alpha}=|\alpha^{\vvv{p}}-\alpha^{\vvv{p}'}|
\end{equation}
The total optical flow loss is given by:
\begin{equation}
     \mathcal{L}_{\textit{flow}}= W^{\vvv{p}} \cdot (\beta_{f}E_{\textit{Flow}}+ \beta_{f\text{-}\alpha}E_{\textit{Flow-}\alpha})
\end{equation}
where $\beta$s are the term weights, and $W^{\vvv{p}}=0$ if the correspondence between $\vvv{p}$ and $\vvv{p}'$ is inconsistent according to standard forward-backward flow consistency check, otherwise $W^{\vvv{p}}=1$,
\update{.} In other words, we mask this loss in regions where the optical flow is unreliable (we use a threshold of 1 pixel).

\paragraph{Sparsity loss.}
The sparsity loss is intended to prevent duplicate representations in the foreground atlas, and further encourage the many-to-one mapping of scene points to atlas points.
Let us consider a pixel $\vvv{p}$;  intuitively, if $\vvv{p}$ is mapped to the background atlas ($\alpha^p \approx 0)$, we know it is not visible in the foreground atlas, and hence should not contain  information about the foreground object in that location. We thus encourage the foreground  atlas at the corresponding location to have black (zero) color.
\begin{equation}
    \mathcal{L}_{\textit{sparsity}} = \beta_{s}\norm{\left(1-\alpha^\vvv{p}\right)\mathbf{c}^\vvv{p}_f}^2 
\end{equation}
where $\mathbf{c}^\vvv{p}_f$ is the predicted color in $\vvv{p}$ \update{for the foreground layer} (see Eq.~\ref{eq:colors}).

\begin{figure*}[t]
\includegraphics[width=.94\linewidth]{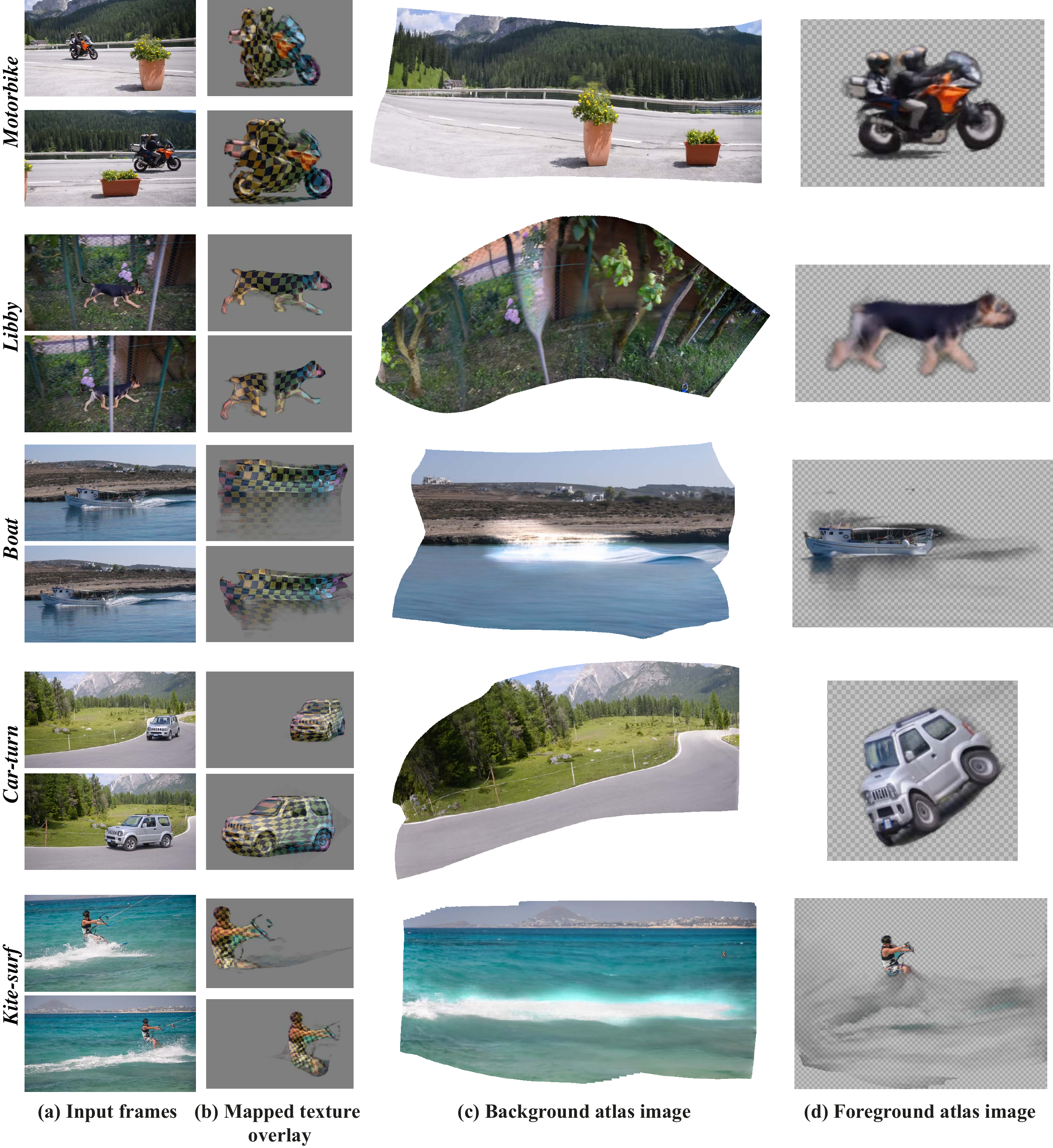}
\caption{\label{fig:qual_results} \textbf{Qualitative results.} Distant input frames (a) from each video sequence are used to create background and foreground atlas visualizations (c,d) as described in Sec.~\ref{sec:results}).
A uniform checkerboard texture was placed on the object region in the foreground atlas and then mapped back to the input frames (b). 
The deformation of the checkerboard pattern reveals the continuity of the atlas mapping over time.}
\end{figure*}

\paragraph{Initialization.} 
We additionally introduce two extra losses that are used \emph{only} during an initial bootstrapping phase (10,000 iterations).
First, to initialize the alpha mapping, we encourage the predicted alpha values to match a coarse input mask that identifies which regions should be split into different layers.
We experienced with two ways to obtain the coarse masks: using a pretrained object segmentation method (MaskRCNN~\cite{he2017mask}), or using a simple polygonal ``garbage'' mask\update{s} provided by the user (See Sec.~\ref{sec:coarsemask}). 
Unless otherwise specified, the results presented are generated using a coarse mask provided by MaskRCNN (similar to~\cite{lu2020layered}).
Let $m^\vvv{p} \in \{0,1\}$ be the foreground label of the sampled video location $\vvv{p}$. 
The bootstrapping loss term is defined by $BCE\left(m^{\vvv{p}},\alpha^{\vvv{p}}\right)$
where \textit{BCE} is the binary cross entropy loss.
We note that we do not require input masks to be precise, as they are used only for initializing the opacity network $\mathbb{M}_\alpha$, and the majority of training happens without this loss, allowing the network to correct for errors.

Second, we initialize with a second \emph{global} rigidity term similar to $\mathcal{L}_{\textit{rigid}}$ but applied to a larger area with $\Delta=100$. The purpose of this term is to initialize the overall structure of the atlas and to prevent distortion of larger objects. 
This term too is turned off after the bootstrapping phase. 

\subsection{Video Editing in Atlas Space}
\label{section:inference}

\update{While our atlases are continuous and thus resolution agnostic}, for editing and visualization purposes, we render a $1000 \times 1000$ discretized version of each atlas. 
The user then performs their desired edits on these discretized atlases.
For example, the user can map textures onto an objects by loading its discretized atlas into image editing software (e.g., Photoshop), and simply drawing the desired texture as a layer on-top of the atlas image.
\update{The edit atlases are then mapped back to the video using the computed UV mapping, and blended with the \emph{original} (not reconstructed) video frames.}

Formally, given a video location $\vvv{p}$, its associated UV coordinates $(u_b^{\vvv{p}}, v_b^{\vvv{p}}),(u_f^{\vvv{p}}, v_f^{\vvv{p}})$ are used to bilinearly \update{sample} the  \update{edits'} RGBA values, which we refer to as $(\vvv{c}_{eb}^\vvv{p},\alpha_{eb}^\vvv{p}),(\vvv{c}_{ef}^\vvv{p},\alpha_{ef}^\vvv{p})$ \update{for the background  and foreground, respectively}.
\update{Edited background and foreground video layers are then formed by  blending the edits' color with the original video color at $\textbf{p}$:}
\begin{equation}\label{eq:edit2}
\begin{array}{c}
    \tilde{\vvv{c}}^{\vvv{p}}_b = (1 - \alpha^{\vvv{p}}_{eb}) \bar{\vvv{c}}_{\vvv{p}} +  \alpha^{\vvv{p}}_{eb} \vvv{c}^{\vvv{p}}_{eb}
\\ \\
\tilde{\vvv{c}}^{\vvv{p}}_f = (1 - \alpha^{\vvv{p}}_{ef}) \bar{\vvv{c}}_{\vvv{p}} +  \alpha^{\vvv{p}}_{ef} \vvv{c}^{\vvv{p}}_{ef}
\end{array}
\end{equation}

\update{Finally, the edited video color for $\vvv{p}$ is given by blending the above background and foreground colors using Eq.~\ref{eq::pixel_reconstruction}}: \begin{equation}\label{eq:edit1}
      \vvv{c}^{\vvv{p}} = (1 - \alpha^{\vvv{p}}) \tilde{\vvv{c}}^{\vvv{p}}_b +  \alpha^{\vvv{p}} \tilde{\vvv{c}}^{\vvv{p}}_f
\end{equation}

Sometimes the user may desire to directly edit input frames, rather than the atlas. 
In this case, we can take the edit \update{of the} frame and project it onto the discretized atlases using $\mathbb{M}_f,\mathbb{M}_b$, and generate the final video as before.

\subsection{Implementation Details}\label{section:implementation_details} 

Please see Appendix~\ref{app:arch} for an exact description of our architectures.
All of our networks use \update{Rectified Linear Unit (ReLU)} activation between hidden layers, and outputs are passed though a tanh function.
This way the model maps the atlases to a pre-defined continuous range.
We use an Adam optimizer with a learning rate of 1e-4 for optimizing all networks simultaneously.

We work with videos consisting of 70 frames with resolution $768\times 432$.
We use a batch size of 10,000 point locations and train each model for around 300,000 iterations. 
\update{In total, our model has around 1.2M learnable parameters, \mytilde 200K for each mapping network, and \mytilde 400K for the alpha and atlas networks.
Training and  evaluation requires \mytilde 5GB GPU memory and the checkpoint size is 14.3MB.}
Training our model takes about 10 hours per video, and reconstructing every video takes roughly 7 minutes \update{on} a\update{n} NVIDIA Quadro RTX 6000 GPU.

Our method has a number of weights for each loss term, which we determine empirically, and fix for all results shown in the paper.
For the reconstruction loss, we set $\beta_r=5000$ and $\beta_g=1000$, for the rigidity term, we set $\beta_r=5$, for the optical flow term we set $\beta_f=5$ and $\beta_{f\text{-}\alpha}=50$, and for the regularization term we set $\beta_{\textit{tv}}=100$ and $\beta_{s}=1000$.

\update{In all of our Jacobian computations, we use finite differences.  Although auto-diff can be used for computing the local rigidity loss (and works similarly to our finite differences implementation), it cannot be used for the global rigidity loss. To have a unified framework, we compute both losses using finite differences.}
\section{Results}
\label{sec:results}

\begin{figure}
	\renewcommand{\arraystretch}{.8}
	\def\myheight{2.0cm}
	\small
	\begin{tabular}{*{4}{c@{\hspace{1px}}}}		
		\includegraphics[clip,trim=125 0 425 0,height=\myheight]{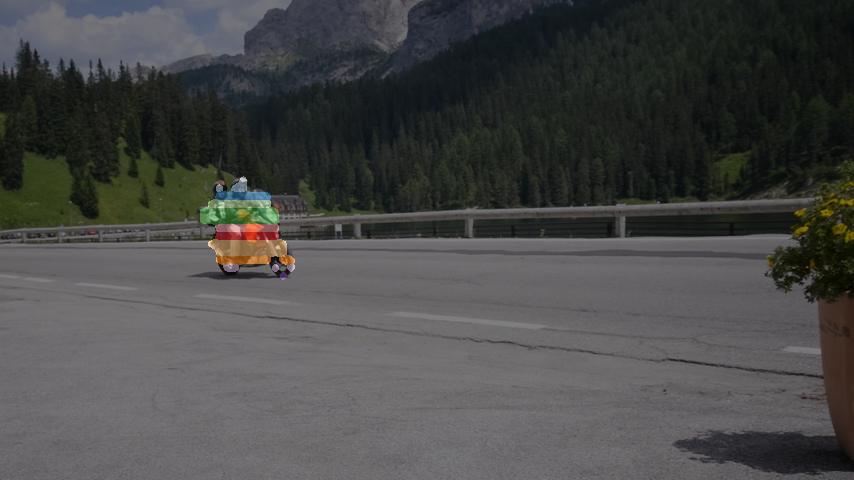} & 
		\includegraphics[clip,trim=125 0 150 0,height=\myheight]{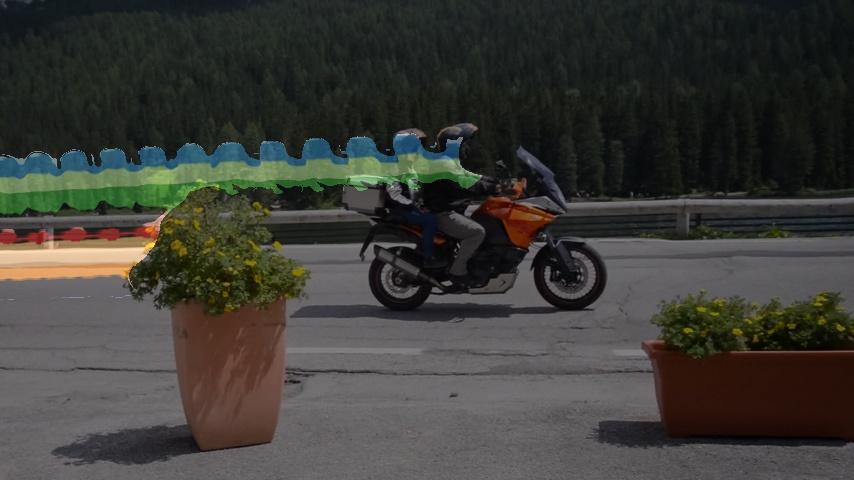} & 
		\includegraphics[clip,trim=125 0 150 0,height=\myheight]{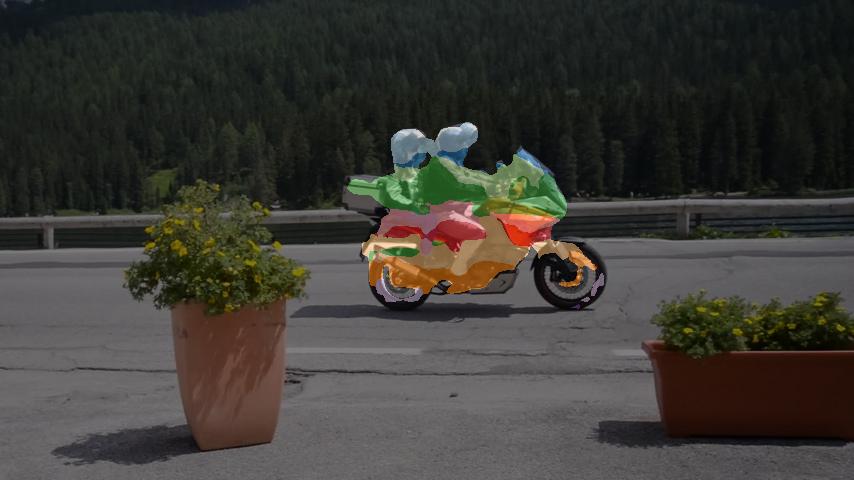} & 
		\includegraphics[clip,trim=125 0 150 0,height=\myheight]{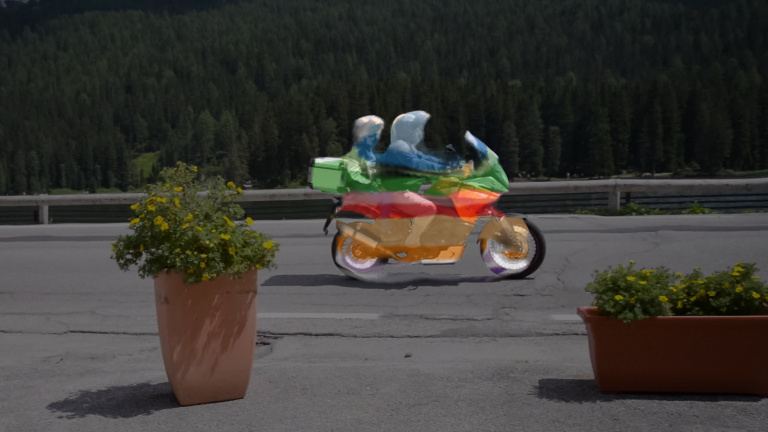}\\
		\includegraphics[clip,trim=275 0 275 0,height=\myheight]{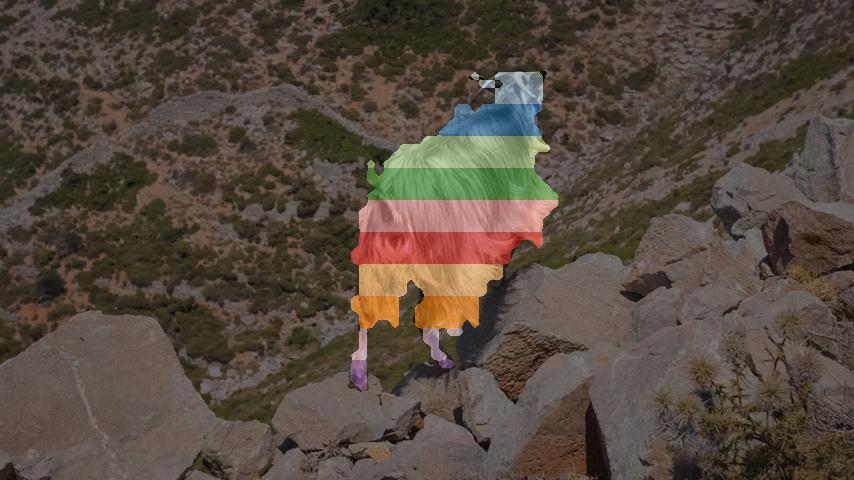} & 
		\includegraphics[clip,trim=125 0 150 0,height=\myheight]{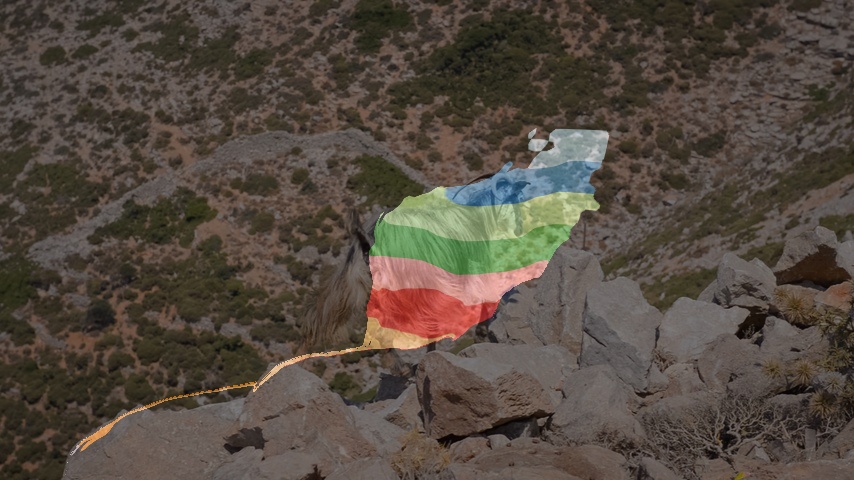} & 
		\includegraphics[clip,trim=125 0 150 0,height=\myheight]{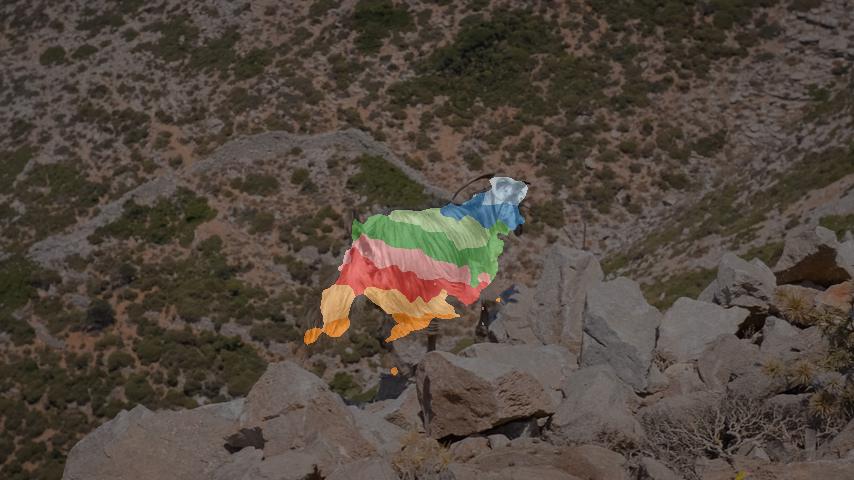} & 
		\includegraphics[clip,trim=125 0 150 0,height=\myheight]{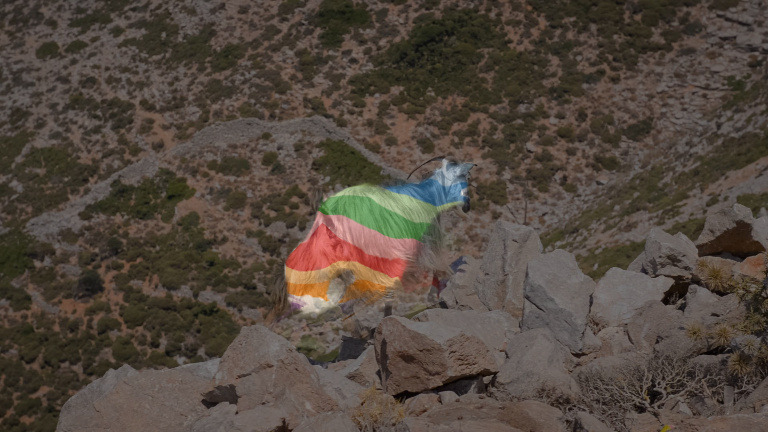} \\
		\includegraphics[clip,trim=125 0 425 0,height=\myheight]{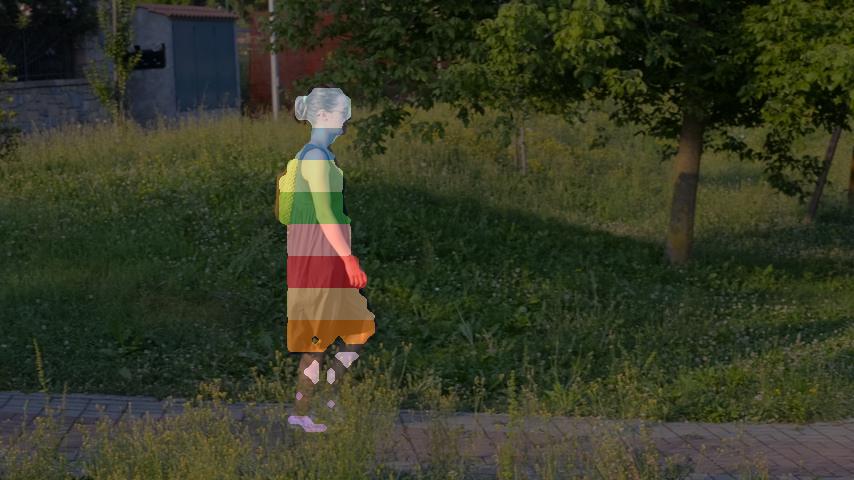} & 
		\includegraphics[clip,trim=125 0 150 0,height=\myheight]{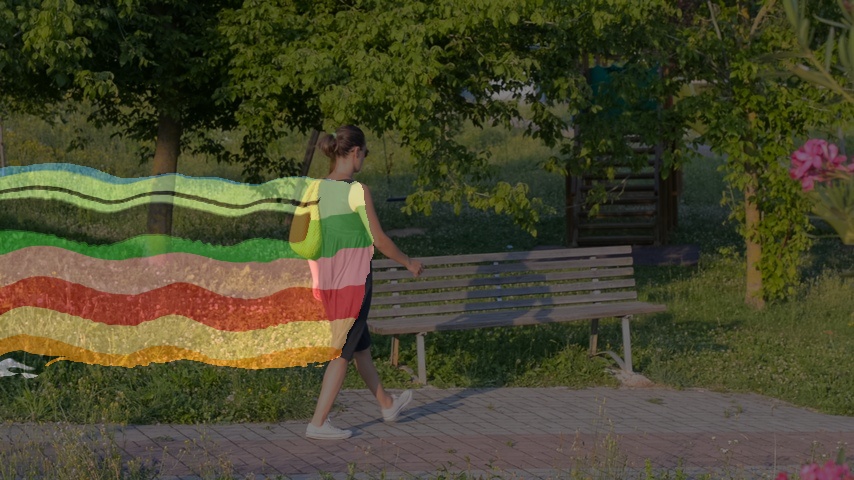} & 
		\includegraphics[clip,trim=125 0 150 0,height=\myheight]{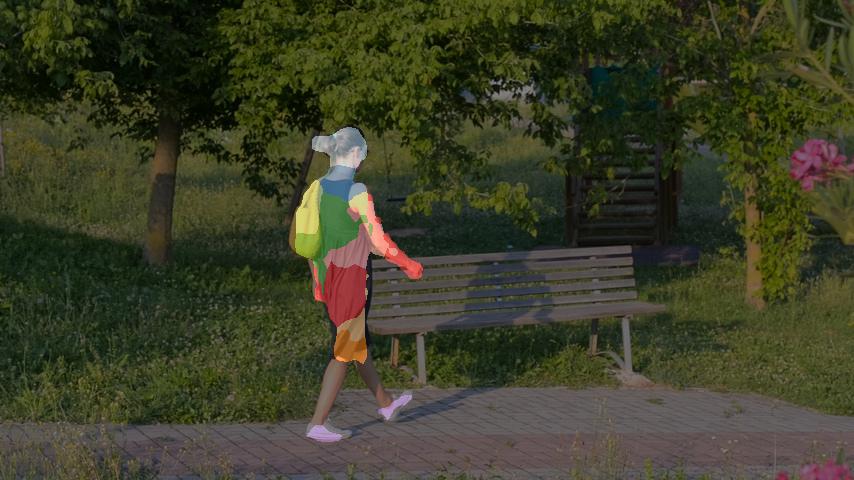} & 
		\includegraphics[clip,trim=125 0 150 0,height=\myheight]{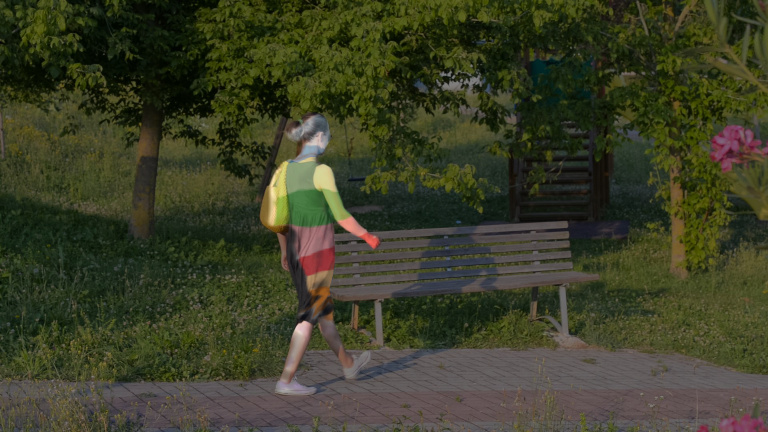} \\
		{\bf Input frame} & {\bf Flow baseline} & {\bf Videowalk~\shortcite{jabri2020space}} & {\bf Ours} \\
	\end{tabular}
	\caption{\label{fig:comparisons}\textbf{Video propagation.} An image with color stripes is given on the first frame (left), and propagated through the video. The flow baseline causes streaking due to ambiguities at motion disocclusions;  Videowalk successfully tracks the object, but results in temporally inconsistent mapping, as observed by zigzagging patterns in the stripes, and \supp{as flickering in the video results in the supplementary material.} Our approach generates a more consistent parameterization.}
\end{figure}

\begin{figure*}[t!]
    \centering
    \includegraphics[width=\linewidth]{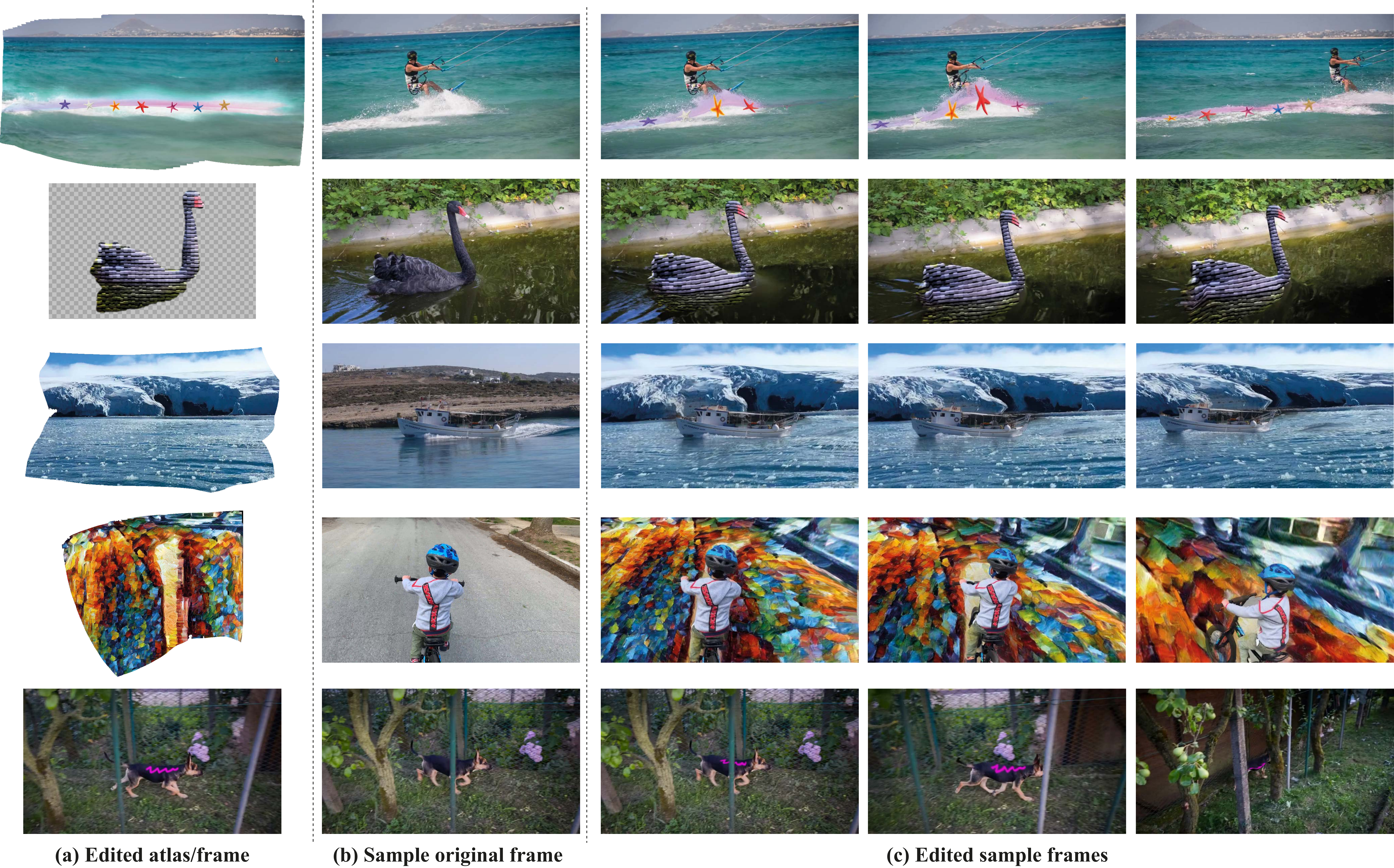}
    \caption{\textbf{Video editing applications.} (a) Various editing effects are  applied directly on our output atlases (top 4) or  on a given video frame (bottom); these effects include stylizing foreground object (\emph{Blackswan}) or background (\emph{Bicycle}), transferring texture elements (\emph{Kite-surf}, \emph{Libby}), or transferring a still image into a moving background (\emph{Boat}). In all cases, edits are automatically and \update{consistently} mapped to the original video frames. \supp{Please see the supplemental material for video examples of these edits.}}
    \label{fig:applications}
\end{figure*}

\begin{figure}[t]
\includegraphics[width=.5\textwidth]{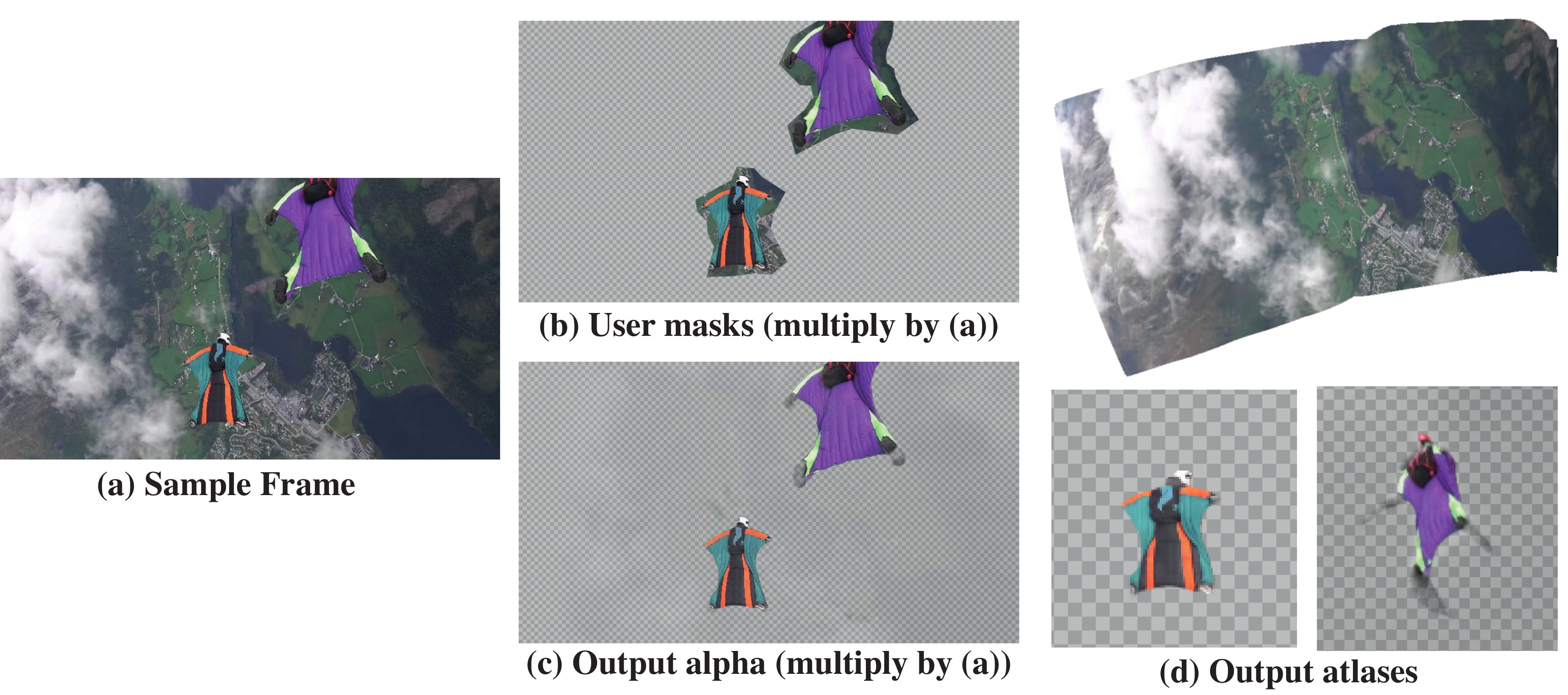}
\caption{\label{fig:coarsemaskmultiple} {\bf Coarse input masks and multiple layers.} When provided with coarse masks by the user (b), the network is still able to generate an accurate layer decomposition with fine detail (c), and atlases (d).\vspace{-2mm}}
\end{figure}

\begin{figure*}[t]
\includegraphics[width=.9\linewidth]{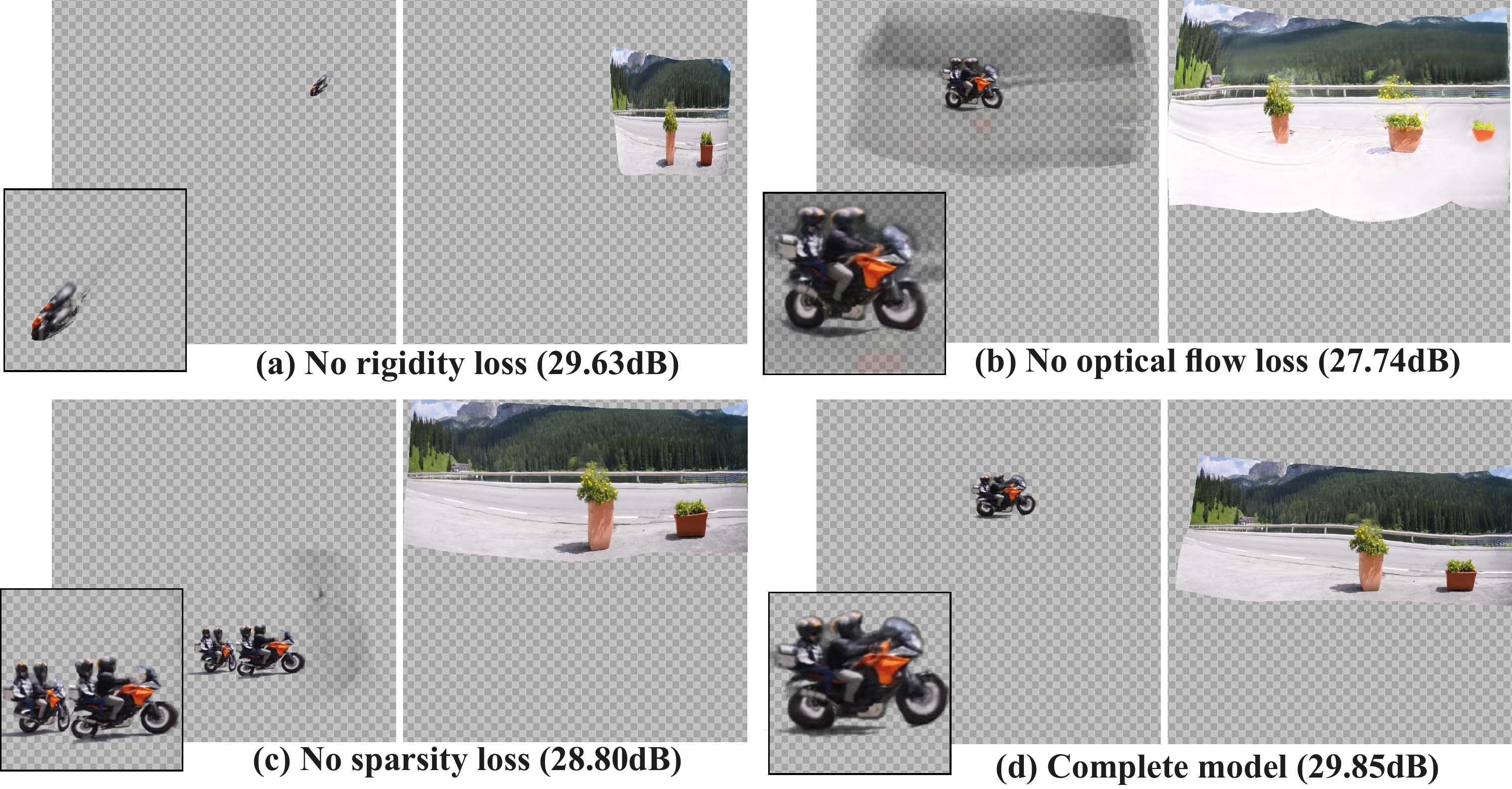}
\caption{\label{fig:ablation} \textbf{Loss ablation.} Visualization of the non-cropped atlases when removing loss parts from total objective loss. The resulting atlases show duplicated objects, distortions and \update{noisier} decomposition into foreground/background layers compared with our complete model. See Sec.~\ref{sec:results} for more details.}
\end{figure*}

We tested our method on videos taken from the DAVIS dataset \cite{davis2017} as well as our own videos, containing various moving objects (different animals, people, cars), many of which contain background parallax. 
The goal of our approach is consistent video editing, and as such there are no reasonable quantitative baselines to provide.
We thus present a qualitative evaluation of our outputs, demonstrate various editing results, perform careful ablation for each of our losses, and compare the consistency of mapping to a state-of-the-art video label propagation method and optical flow baseline.
All of our results are included in video form in the SM, and we refer the reader to Appendix~\ref{app:psnr} for the reconstruction Peak Signal-to-Noise Ratio (PSNR) values.

\paragraph{Atlas visualization.}
Our representation uses continuous \emph{time-varying} opacity values, and \emph{time-invariant} atlas colors.
However, for visualization purposes, it is useful to see which parts of the atlas are actually used. 
For this, we need to assign a single alpha value to the atlas. 
We do this by discretizing the $\mathbb{M}_\alpha$ onto the same 1000x1000 grid as the atlas, and take the median of all alpha values for the foreground atlas, and the maximum of all alphas for the background atlas. 
This visualizes the entire background, and the most common foreground regions. 
It is important to note that atlases shown in figures are only a \emph{visual proxy} of our reconstruction, and as a result, the visualization may not represent exactly the subtle time varying semi-transparent elements.

In Fig.~\ref{fig:qual_results} we show several qualitative visualizations of our atlases on videos from the DAVIS dataset. 
For each example, we visualize the background and foreground atlas layers using the method described above.
In each example we can see that our method reconstructs interpretable atlases for both background and foreground regions.
In addition, we can see that our parameterization is temporally consistent in that the atlases lack duplication of features, and the colored checkerboard patterns stay fixed on the surface of the objects across time in Fig.~\ref{fig:qual_results}(b).

\paragraph{Video editing applications.}
Fig.~\ref{fig:applications} and Fig.~\ref{fig:teaser} show several video editing results.
As can be seen, our method supports a variety of video effects including transferring texture (\emph{Kite-surf} and \emph{Libby}), stylizing the background or foreground (\emph{Blackswan} and \emph{Bicycle}), or transferring a still image into a moving background in a video (\emph{Boat}).
We also observe that our method preserves complex effects such as reflections, shadows and occlusions.
Please see the supplemental video results. 

In all cases, edits were made in a matter of minutes by non-professional users.
For texture transfer, the user loads the atlas in an editing software such as Photoshop (PS), and adds texture elements by directly drawing with a brush tool or placing a desired image overlay on the atlas.
We also show an example of editing a frame directly, and projecting this onto the atlas and then back into the video in \emph{Libby}.
In \emph{Blackswan}, we applied an artistic filter (\emph{Stained Glass} in PS) to a selected region in the foreground atlas.
In \emph{Bicycle}, we applied an existing image style-transfer method~\cite{kolkin2019style} directly to our background atlas.

\subsection{Comparison to Prior Work}

We compare the correspondences derived by our approach to those from a recent state-of-the-art label propagation method~\cite{jabri2020space}, and our own flow-based baseline.

\paragraph{Comparison to video propagation methods.} 

Another task that requires computing a consistent mapping of content from one frame to the rest of a video is label propagation.
Our method differs slightly in that the goal of video propagation is to label \emph{all} pixels corresponding to the same object class, whereas our goal is to propagate the input \emph{only} to the specific parts of the object that are labeled in the first frame.
Nonetheless, we compare correspondences over time to a recent state-of-the-art self-supervised video propagation approach~\cite{jabri2020space}, called \textsc{Videowalk}.
We run the authors code with a label map consisting of horizontal colored stripes on the first frame of the video as input, and use the same input for our approach. 

Second, we implement an optical flow propagation baseline that accumulates optical flow fields computed between consecutive frames, and uses the accumulated flow to warp the edit into each target frame (\textsc{Flow baseline}).

Fig.~\ref{fig:comparisons} shows a comparison between our approach and these two baselines. 
We can see that \textsc{Flow baseline} produces good temporal continuity where the flow can be chained accurately, however it breaks at occlusions.
In addition, motion disocclusions cause errors at the borders of objects (e.g, the streaking effect in Fig.~\ref{fig:comparisons}).
\textsc{Videowalk} accurately labels the entire object in each frame, but the results exhibit temporal instability, which can be seen by the distortion of the input horizontal stripes in the later frames, and in the supplementary video.
Our approach produces better results in terms of temporal consistency.

\begin{figure}[t]
    \includegraphics[width=\linewidth]{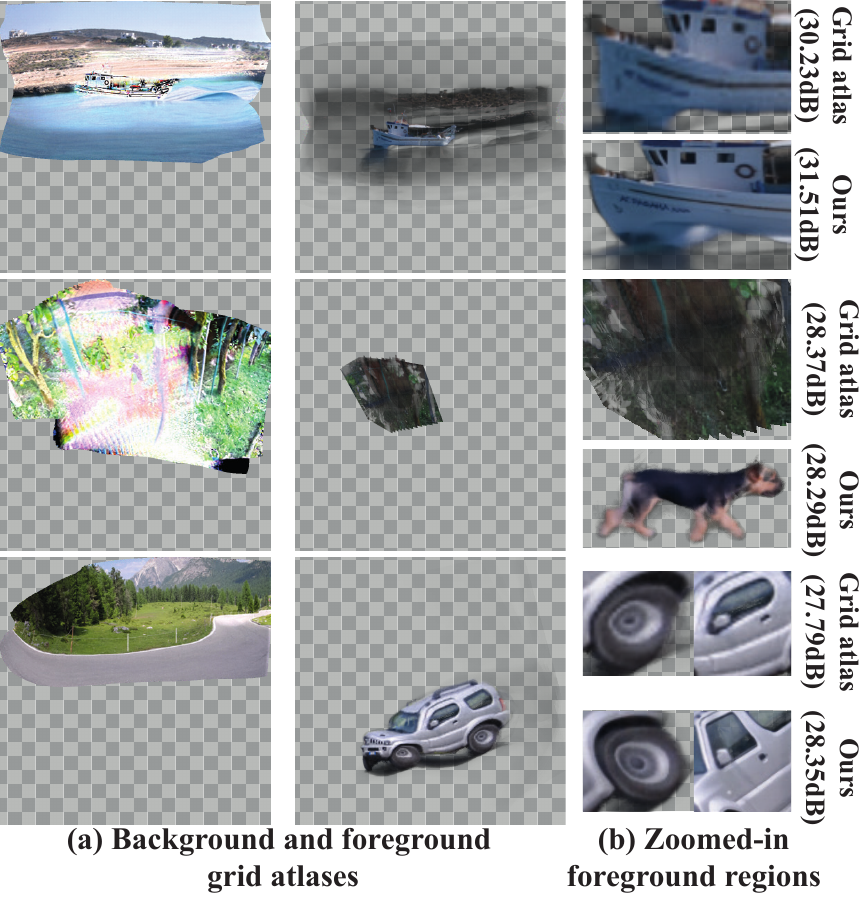}
    \caption{\textbf{Architecture ablation.} When we compare our coordinate-based MLP architecture to a discrete version (\textsc{Grid atlas}), we can see that our method reconstructs higher fidelity textures. In some cases, such as the dog example, \textsc{Grid atlas} fails all together to reconstruct the foreground object. \update{PSNR is computed between the reconstruction and original input video.}}
    \label{fig:gridatlas}
\end{figure}

\subsection{Ablation Study}
\label{sec:ablation_res}
We validate the choice of losses for our model with an ablation study, visualized in Fig.~\ref{fig:ablation}, where we compare the atlases reconstructed by our complete model, and our model with a specific loss removed. 
As can be seen, without the \textbf{rigidity loss}, both atlases become distorted to the point where the foreground object cannot easily be recognized (Fig.~\ref{fig:ablation}(a)), and intuitive editing is not possible.
Without the \textbf{optical flow} loss, the foreground object and backgrounds are reasonably reconstructed, however we notice that the larger plant is duplicated and split (Fig.~\ref{fig:ablation}(b)); furthermore, the decomposition is less constrained, resulting in background elements in the foreground layer.  Somewhat interestingly, the reconstruction is actually reasonable (there is only minimal duplication), which means that the network is learning correspondences to re-use atlas content even without the optical flow loss.
Finally, without the \textbf{sparsity loss}, we see that the entire foreground object has been duplicated in the atlas (Fig.~\ref{fig:ablation}(c)).
This is because the network can improve the reconstruction by simply storing multiple copies of the object, and in this case as the object is occluded, it does not have to connect the two instances together. 
Editing such an atlas would be difficult as the same edit would have to be applied to both duplicated objects. 
We note that the reconstruction quality measured using PSNR is comparable under all configurations. Importantly, even though  our complete model is more constrained, the overall reconstruction quality is persevered. 

\paragraph{Continuous representation.}
\label{sec:gridatlas}
We ablate the choice of using an MLP for representing the atlases vs traditional image-based representations.
In this experiment, we replace our atlas MLP with a fixed resolution (1000x1000) texture image, and optimized it directly with the rest of our components (mapping and alpha networks) using the same objective.
We refer to methods trained with this approach as \textsc{Grid atlas}. Fig.~\ref{fig:gridatlas} shows a comparison of the results. \update{We observe that for complex videos, such as \emph{Libby} and \emph{Boat} in Fig. \ref{fig:gridatlas}, the optimization using the discrete \textsc{Grid atlas} does not converge into a useful decomposition of the scene.
We believe the reason is twofold: (1) Solving first order optimization problems in the space of the network’s weights implicitly imposes regularizations on the optimization problem, which favor “plausible/natural” solutions. This is due to the inductive and spectral bias of neural networks \cite{ulyanov2018deep,arora2019fine}. (2) Using the discrete atlas requires correctly allocating the resolution and extent (in pixels) of regions in the atlas space, before you actually know the shape or scale of the atlas.
Incorrectly discretizing your atlas space in early stages of the optimization can lead to wrong results and pixelated atlases. In contrast, our coordinate-based MLP atlas is easy to use as it is fully differentiable without requiring sampling or interpolation, and leads to cleaner and high quality atlases.}

\update{We note that even in cases where the decomposition of the \textsc{Grid atlas} model is poor, the reconstructed video may still look reasonable (and have a high PSNR).
This is because the network is able to ``cheat'' by drawing the foreground object using the alpha channel, grabbing the desired colors from the background or foreground atlases.
Nevertheless, in such cases, the mapping is inconsistent and distorted, and thus these regions of the video are not editable. 
An example is presented in the SM.}

\subsection{Multiple Foreground Atlases}
\label{sec:multiple_object}
\label{sec:coarsemask}
Our model can support multiple layers with minor modifications.
Let $i$ be the index of the $i^{th}$ foreground layer out of $n$ layers in total, we denote its mapping network by $\mathbb{M}_{f_i}: \mathbb{R}^3\rightarrow \mathbb{R}^2$.
In this setup $\mathbb{M}_\alpha$ produces $\vvv{\alpha}^\vvv{p}\in \mathbb{R}^{n+1}$ outputs followed by Softmax activation, satisfying: $\sum_{i=0}^{n}\alpha^\vvv{p}_i=1$. 
$\mathbb{M}_{f_b}, \mathbb{M}_\alpha, \mathbb{A}$ are defined as in Sec.\ref{section:training}.
Then, a pixel location $\vvv{p}$ is reconstructed by: 
\begin{equation}
     \vvv{c}^{\vvv{p}} = \alpha_0^{\vvv{p}} \vvv{c}^{\vvv{p}}_b +  \sum_{i=1}^{n}\alpha_i^{\vvv{p}} \vvv{c}^{\vvv{p}}_{f_i}
\end{equation}
where,
\begin{equation}
    \begin{array}{c}
    (u_b^{\vvv{p}}, v_b^{\vvv{p}}) = \mathbb{M}_b(\vvv{p}),(u_{f_i}^{\vvv{p}}, v_{f_i}^{\vvv{p}}) = \mathbb{M}_{f_i}(\vvv{p}) \\ \\ \vvv{c}_b^{\vvv{p}} = \mathbb{A}(u_b^{\vvv{p}}, v_b^{\vvv{p}})
    ,\vvv{c}_{f_i}^{\vvv{p}} = \mathbb{A}(u_{f_i}^{\vvv{p}}, v_{f_i}^{\vvv{p}})
    \end{array}
\end{equation}
Unlike the one foreground object case, to support occlusions between different foreground objects, the sparsity loss is applied directly on the atlas, by applying $l_1$ regularization on randomly sampled UV coordinates from foreground regions in the atlas. 
We show an example of a reconstruction using multiple object atlases in Fig.~\ref{fig:coarsemaskmultiple}.
This example also demonstrates the use of a coarse ``garbage'' mask provided by a user in the form of a rough polygon, tracked over time.
The network successfully refines this mask, generating an accurate layer decomposition, as seen in Fig.~\ref{fig:coarsemaskmultiple}(d), while assigning each object to its own atlas. 

\subsection{User Interaction}\label{sec:user_input}

\begin{figure}[t]
\includegraphics[width=.4\textwidth]{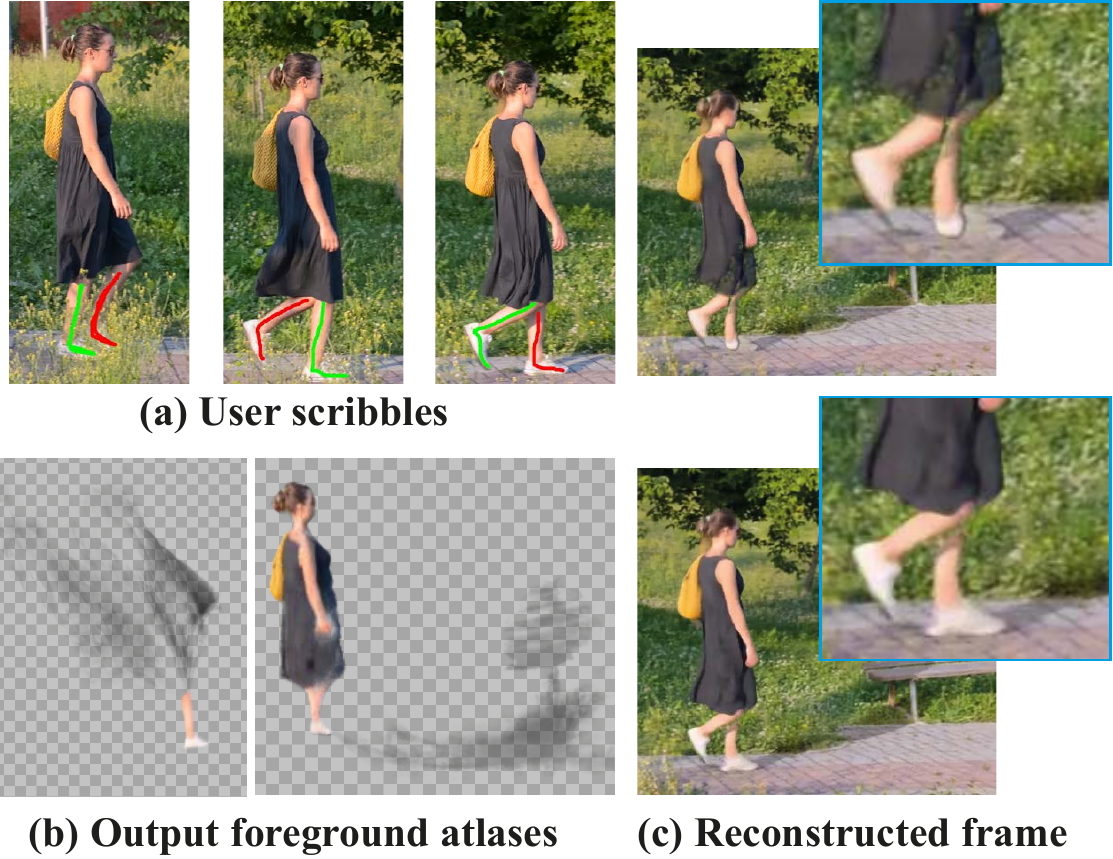}
\caption{\label{fig:scribbles}\textbf{User scribbles}. The user adds red and green scribbles (a) indicating separation into two different foreground atlas layers, i.e., the green scribbled pixels should map to one foreground layer and the red pixels should map to another foreground layer. This separation allows the model to correctly represent each leg despite self-occlusion and complex deformations (b), which leads to better reconstruction of a sample frame (c). See SM for the editing result w/ and w/o user scribbles. \vspace{-2mm}}
\end{figure}

For videos containing one foreground object with self occluded areas (e.g. the legs of a walking person),  it is difficult for one foreground layer to represent self occluded parts correctly. To address this, our method can optionally use rough scribble inputs from the user for determining the assignment of occluded regions of one foreground object to two different layers.
The user input can be simple, e.g., brush curve with different colors indicating the two layers (see Fig.~\ref{fig:scribbles}(a)).
The foreground layer separation is only constrained by the sparse set of user scribbles during the bootstrapping phase. Specifically, we change our mask loss to be applied on the summation of the two foreground layers: \begin{equation*}
    BCE(m^p,\alpha_1^{\vvv{p}}+\alpha_2^{\vvv{p}})
\end{equation*}  
 In addition, we add the following loss terms:
\begin{equation*}
\begin{array}{c}
     \mathcal{L}_{\textit{Red}}=-\log(\alpha_1^{\vvv{p}_r})
     \\ 
     \mathcal{L}_{\textit{Green}}=-\log(\alpha_2^{\vvv{p}_g})
\end{array}
\end{equation*} 
where  $p_g$ and $p_r$ are the pixels that are are marked as "green" and "red" scribbles respectively.  Fig.~\ref{fig:scribbles} shows our results when  using scribble input to handle self-occlusion in the \emph{Lucia} example.
All other results in the paper do \emph{not} use user scribbles as input. 

\begin{figure}[t]
\includegraphics[width=\linewidth]{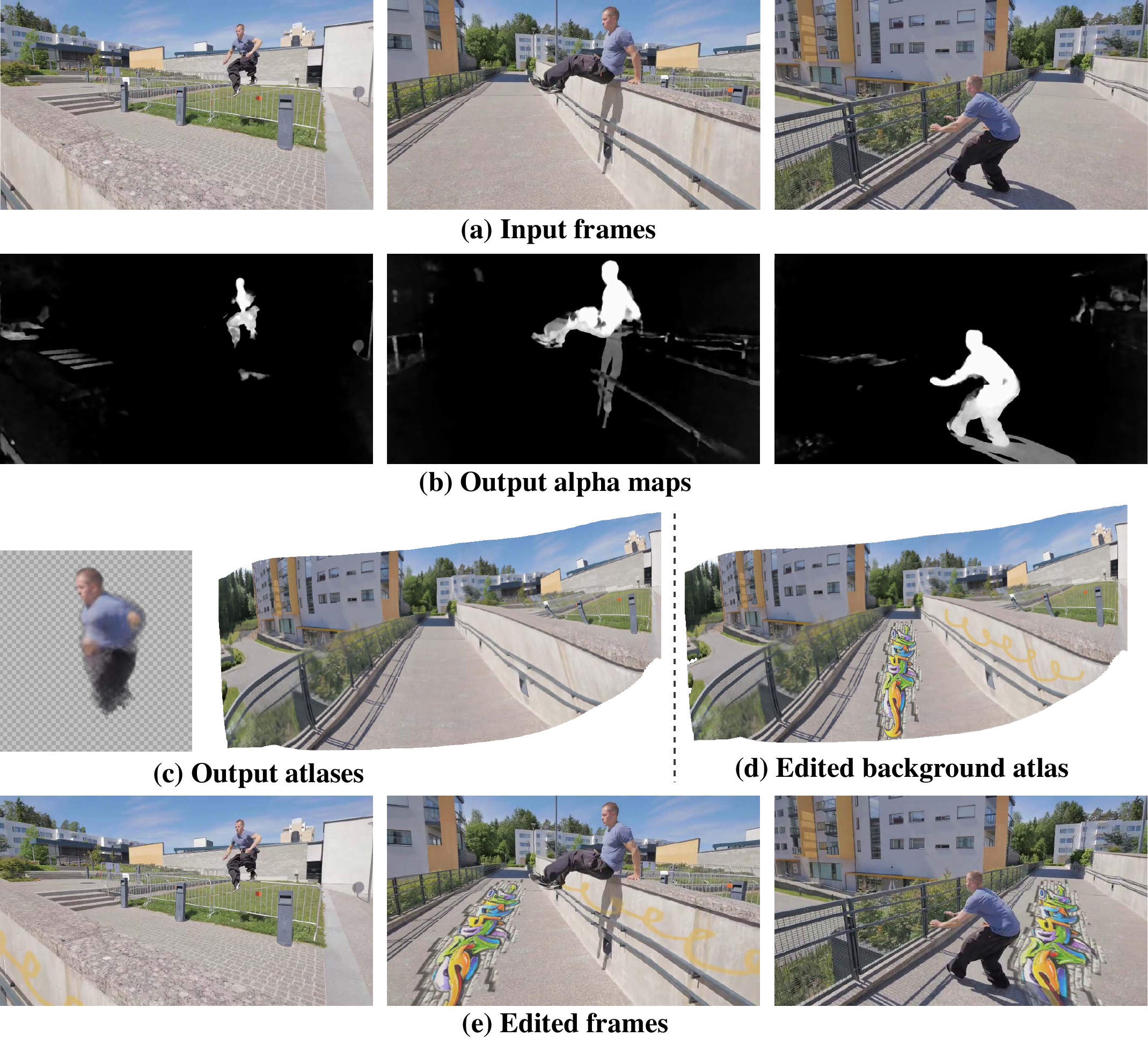}
\caption{{\bf Limitations.} (a) The input video is challenging in that person is observed from both the front and back, while performing a complex action with many self-occlusions, and the background motion contains significant parallax due to camera motion. Our method was not able to accurately represent the person -- the foreground atlas is distorted and missing limbs (b, left). However, the background reconstruction is reasonable and thus supports edits (e).}
\label{fig:faliure}
\vspace{-5mm}
\end{figure}

\subsection{Limitations}
Our approach has a few limitations: our atlases are \emph{fixed in time}, i.e., we do not model temporal appearance changes on the atlases.
Therefore, our framework can model appearance variation in the video (e.g., lighting changes) only through the per-frame alpha-maps and mapping.
This can result in some noise in the foreground atlases in these cases (see \emph{Libby} video example in the SM).
Due to the capacity of our MLP networks, the quality of our results decreases for long videos ($>100$ frames).
In addition, as reported in \cite{lu2021omnimatte}, we also observed that different random initializations of the network’s weights may occasionally result in different solutions; we speculate that a more intelligent batch sampling scheme can improve this issue, yet it requires careful future analysis. 
Similarly to \cite{rav2008unwrap}, we are also limited to disk topology objects, i.e., cannot represent an object repeatedly rotating in space.
Finally, as discussed, it is difficult for the model to faithfully represent complex geometry, self-occlusions and extreme deformations with a single foreground layer.
This can be seen by the missing limbs of the person in  Fig.~\ref{fig:faliure}.
However, we observed that even is such cases our method is still able to correctly decompose the video and faithfully represent the other parts of the scene. We can thus edit those regions and consistently map them to the original video frames (see Fig.~\ref{fig:faliure}(e)).

\section{Conclusion}
In conclusion, we have introduced the first approach for neural video unwrapping using an end-to-end optimized atlas-based representation.
We believe that the frame/atlas-based editing framework is both accessible for beginners and provides a high level of creative control for professionals.
Unlike prior work, our method can be used on a wide variety of input videos, and we demonstrate the advantages of such an approach on a number of useful downstream applications, enabling simple, temporally consistent video editing.
We believe that the idea of decomposing videos through a self-supervised, “interpretable bottleneck”, in our case a set of 2D atlases, has the potential to simplify other complicated video processing tasks.

\section*{Acknowledgments}
We would like to thank Ronen Basri and Noam Aigerman for insightful discussions. This work has been funded in part by the Israeli Science Foundation (grant 2303/20), by the U.S.-Israel Binational Science Foundation, grant No.~2018680 and by the D.~Dan and Betty Kahn Foundation

%
%
%
%


\bibliographystyle{ACM-Reference-Format}
\bibliography{unwrapping}

\appendix
\section{Network Architecture}
\label{app:arch}
The background mapping MLP $\mathbb{M}_b$ has 4 layers with 256 channels each and the foreground mapping MLP $\mathbb{M}_f$ has 6 layers with 256 channels each. 
The alpha MLP $\mathbb{M}_\alpha$ and atlas MLP $\mathbb{A}$ have 8 layers, each with 256 channels. 
We add skip connections in the atlas MLP for layers 4 and 7, concatenating the network input to the features at those layers' inputs, similar to the architecture used in NeRFs~\cite{mildenhall2020nerf}.
To allow the networks to learn to represent higher frequency functions, we use positional encoding for the alpha and atlas MLPs, with number of frequencies $N=5$ and $N=10$ respectively. 

The mapping networks, $\mathbb{M}_b$ and $\mathbb{M}_f$, have a short initial training phase, which serves as a soft ``calibration''. 
We train the identity mapping ($(x,y,t) \rightarrow (x,y)$) for 100 iterations, which helps prevent the mapping from being flipped, and retain the order of points.

\section{Reconstruction PSNR}
\label{app:psnr}
\begin{table}[th]
\setlength\tabcolsep{0pt} 
\begin{tabular*}{0.3\textwidth}{p{4.5cm} p{3cm}}
\toprule
\textbf{Video Name} & \textbf{PSNR} \\
\midrule
Bicycle    & 28.56 \\
Blackswan       & 29.92 \\
Boat            & 31.51 \\
Car-turn        & 28.35 \\
Kite-surf       & 28.37 \\
Libby           & 28.29 \\
Motorbike       & 29.85 \\
Lucia (2 layers)          & 27.43    \\
Parachuting (2 layers)     & 27.24 \\
\bottomrule
\end{tabular*}
\caption{PSNR results for video reconstruction}
\label{tab:psnr}
\end{table}

Tab.~\ref{tab:psnr} shows the reconstruction quality for different videos using our two-atlas approach. 

\end{document}